\documentclass[runningheads]{llncs}
\usepackage{graphicx}
\usepackage{amsmath,amssymb} % define this before the line numbering.
\usepackage{color}
\usepackage[width=122mm,left=12mm,paperwidth=146mm,height=193mm,top=12mm,paperheight=217mm]{geometry}

\usepackage[tight,footnotesize]{subfigure}
\usepackage{rotating}
\usepackage{graphicx}
\usepackage{amsmath,amssymb} % define this before the line numbering.
\usepackage{epsfig}
\usepackage{array}
\usepackage{multirow}
\usepackage{colortbl}
\usepackage{amsfonts}
\usepackage{pifont}
\usepackage{xspace}
\usepackage{footnote}
\usepackage{etoolbox}
\usepackage{cite}
\usepackage{overpic}
\usepackage{epstopdf}

\begin{document}
% \linenumbers
%\pagestyle{headings}
\mainmatter
\def\ECCV16SubNumber{779}  % Insert your submission number here

\title{Deeply-Fused Nets} % Replace with your title

%\titlerunning{ECCV-16 submission ID \ECCV16SubNumber}

%\authorrunning{ECCV-16 submission ID \ECCV16SubNumber}

\author{Jingdong Wang\textsuperscript{1}, Zhen Wei\textsuperscript{2}, Ting Zhang\textsuperscript{3}, Wenjun Zeng\textsuperscript{1}}
\institute{\textsuperscript{1}Microsoft Research, \textsuperscript{2}Shanghai Jiao Tong University \\ \textsuperscript{3}University of Science and Technology of China}

\maketitle

\begin{abstract}
In this paper,
we present a novel deep learning approach,
deeply-fused nets.
The central idea of our approach is deep fusion,
i.e., combine the intermediate representations of base networks,
where the fused output serves as the input of the remaining part
of each base network,
and perform such combinations deeply
over several intermediate representations.
The resulting deeply fused net enjoys several benefits.
First, it is able to learn multi-scale representations
as it enjoys the benefits of more base networks,
which could form the same fused network, other than the initial group of base networks.
Second,
in our suggested fused net
formed by one deep and one shallow base networks,
the flows of the information from the earlier intermediate layer of the deep base network
to the output
and from the input to the later intermediate layer of the deep base network
are both improved.
Last,
the deep and shallow base networks are jointly learnt
and can benefit from each other.
More interestingly,
the essential depth of a fused net
composed from a deep base network
and a shallow base network
is reduced
because the fused net could
be composed from
a less deep base network,
and thus training the fused net
is less difficult than training the initial deep base network.
Empirical results
demonstrate that our approach
achieves superior performance over
two closely-related methods,
ResNet and Highway,
and competitive performance
compared to the state-of-the-arts.

\end{abstract}

\section{Introduction}
Deep neural network has been popular again
since the breakthrough performance~\cite{KrizhevskySH12}
in the ImageNet classification~\cite{DengDSLL009}.
In the past few years (from $2012$),
the top-$5$ classification accuracy on the $1000$-class
ImageNet dataset has increased from $\sim 84\%$~\cite{KrizhevskySH12} to $\sim97\%$~\cite{SzegedyIV16}.
Besides,
deep neural network has been shown to have very impressive performance
for other vision applications,
such as object detection~\cite{GirshickDDM14, GirshickDDM16}, image segmentation~\cite{LongSD15},
edge detection~\cite{XieT15}, and so on.

Nevertheless,
the fundamental problem,
learning a deep hierarchical structure effectively and efficiently,
still remains a challenge
and has been attracting a lot of research efforts.
Dropout~\cite{SrivastavaHKSS14} and other regularization techniques,
such as
weight decay and path regularization~\cite{NeyshaburSS15},
have been developed
to prevent neural network from over-fitting.
Normalized variance-preserving weight initialization,
such as~\cite{GlorotB10, HeZRS15a, MishkinM15},
has been shown to be helpful for handling the vanishing gradient problem
and thus boosts the performance.
Batch normalization~\cite{IoffeS15} is shown to improve
both the training speed and the recognition performance.
Skip-layer connections between layers (including the output layer)
and other network structure modifications,
such as
deeply-supervised nets~\cite{LeeXGZT15} and its variant~\cite{XieT15},
Highway~\cite{SrivastavaGS15a},
ResNet~\cite{HeZRS15b}, inception-v$4$~\cite{SzegedyIV16},
are able to improve the flow of information
and accordingly help train a very deep network.
The teacher-student framework shows that learning a deep network
can benefit from an already-trained network
that is relatively easy to be learnt, e.g., FitNets~\cite{RomeroBKCGB14} and Net2Net~\cite{ChenGS15}.

In this paper,
we introduce a deep fusion approach and present a deeply-fused neural net formed by combining a group of base networks.
The main idea is
to perform fusion over the intermediate representations of the base networks,
where the fused output serves as the input of the remaining part of each base network,
rather than only over the final representations or the final classification scores,
and such fusions are performed several times at different intermediate layers.
There is a block-exchangeable property
(the block is the subnetwork between two successive fusions in a base network):
switch blocks from one base network
to another one within one fusion,
resulting in two different base networks
with possibly different depth from the originals (e.g., deep network being less deep and shallow network being less shallow),
but the fused net is not changed.
In other words,
a fused net
can be formed by different groups of base networks.
Thus,
the deeply-fused net
is able to learn multi-scale representations from much more base networks,
and even same-scale representations can be different and learnt
from different base networks.

There is one more benefit from deep fusion:
the flow of information is improved.
Consider the case where one base network is very deep but the other base network is not deep,
which is the choice we suggest.
The earlier intermediate layer in the deeper base network might have a shorter path
through the other base network
to the output,
which implies that the supervision can be fast transformed to the earlier intermediate layer.
On the other hand, the later intermediate layer might also have a shorter path from the input,
which indicates that the input can be fast flowed to the later intermediate layer.
As a result,
training the fused net composed from a very deep base network
is less difficult than training the very deep base network itself.
Furthermore,
the deep and shallow base networks are jointly learnt
and can benefit from each other.
We also show that the recently-developed networks, Highway and ResNet,
can be viewed as specific examples of deep fusion.
Empirical results
demonstrate that our approach
achieves superior performance over
the plain network,
the naive network fusion method,
ResNet and Highway,
and competitive performance
compared to the state-of-the-arts.

%In addition,
%there are two more benefits from deep fusion.
%One the one hand,
%we show that the vanishing gradient problem is alleviated
%when one base network is not deep and even the other base networks are very deep.
%which is the choice we suggest.
%The simple analysis suggests that
%too many fusions is not very helpful
%for alleviating the gradient vanishing problem.
%On the other hand,
%the depth of the fused network
%is equal to the largest
%of all the possible base networks
%and accordingly might be deeper
%than the initial group of base networks.
%Thus there is a potential for the deeper network
%to achieve superior performance.
%Experimental results show
%that deeply-fused net achieves state-of-the-art results over
%the popular datasets.

\section{Related Work}
The past few years have witnessed
the rapid and great progress of deep neural networks
in various aspects, from optimization techniques as well as initialization,
regularization, activation and pooling functions,
network structure design,
to applications.
In this section, we mainly discuss
two closely-related lines:
network structure design
and network optimization with the aid of another already-trained network.

Averaging over a set of network predictors, which we call decision fusion,
is able to improve the generalization accuracy
and has been widely used,
e.g., to boost the ImageNet recognition performance~\cite{KrizhevskySH12, SimonyanZ14a, SzegedyLJSRAEVR15, HeZRS15b}.
Multi-column deep neural networks~\cite{CiresanMS12} presents an empirical study about decision fusion,
later extended to an adaptive version, weighted averaging with the weights depending on the input~\cite{AgostinelliAL13}.
The averaging approach learns each network separately,
which is equivalent to learn the network jointly
that averages the loss functions.
Our approach, in contrast, performs the feature fusion deeply over several intermediate layers
and simultaneously learns the representations of the (base) networks.

The inception module in GoogLeNet~\cite{SzegedyLJSRAEVR15}
can be viewed as
a fusion stage: concatenate the outputs of several subnetworks with different lengths.
It is different from our approach using the summation for  fusion.
The GoogLeNet architecture, consisting of  a sequence of inception modules,
is also a kind of deep fusion, i.e., deep concatenation fusion.
But it is not as direct as our deep summation fusion.
The output of each subnetwork in an inception module
is narrower than the input of the subsequent inception module.
Hence it is necessary to append many channels with all $0$ entries in the output
to match the size with the input of the subsequent inception module
or add more convolution operations to form the fused network.
Skip-layer connection, such as deeply-supervised nets~\cite{LeeXGZT15} and its variant~\cite{XieT15},
Highway~\cite{SrivastavaGS15a},
ResNet~\cite{HeZRS15b}, as we will show, resembles our approach
and can be regarded as special examples of our approach.

The teacher-student framework suggests
that learning a hard-trained network
can benefit from an easily-trained network.
For instance, FitNets~\cite{RomeroBKCGB14} uses the intermediate representation
of a wider and shallower (but still deep) teacher net that is relatively easy to be trained,
as the target of the intermediate representation
of a thinner and deeper student net.
Net2Net~\cite{ChenGS15} also uses a teacher net to help train a (wider or deeper) student net,
through a function-preserving transform
to initialize the parameters of the student net
according to the parameters of the teacher net.
Our approach, in our suggested choice:
including one deep base network
and one shallow (but could still be deep) network,
also uses the shallow network to help train the deep base network,
meanwhile the deep base network also helps train the shallow network,
i.e., they benefit from each other and are trained simultaneously.

\section{Deeply-Fused Nets}
\definecolor{lightblue}{rgb}{0.68, 0.85, 0.9}
\definecolor{lightgreen}{rgb}{0.56, 0.93, 0.56}
\definecolor{pastelviolet}{rgb}{0.8, 0.6, 0.79}

\begin{figure}[t]
\centering
\footnotesize
(a)~{\includegraphics[width=.45\linewidth]{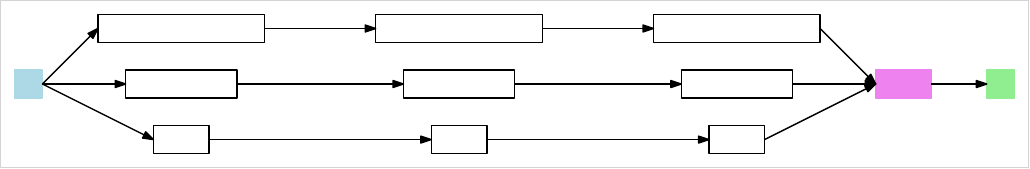}}
(b)~{\includegraphics[width=.45\linewidth]{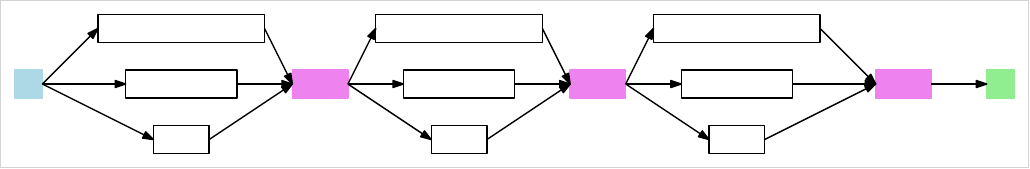}}
\caption{Illustrating (a) shallow fusion and (b) deep fusion.
The {\color{lightblue}light blue}, white,
{\color{pastelviolet}violet},
{\color{lightgreen}light green} boxes correspond to the input,
a block of layers (subnetwork),
the fusion layer,
and the classification layer. (Best viewed in color.)}
\label{fig:compareshallowanddeep}
  \vspace{-.5cm}
\end{figure}

A feedforward network typically
consists of $L$ representation extraction layers, $H_L$,
followed by a classification layer, e.g., a fully-connected layer and a linear classifier.
The $l$th layer applies a nonlinear transformation $h_l$
(parameterized by $\mathbf{w}_l$),
e.g., a linear convolution followed by a nonlinear activation function:
\begin{align}
\mathbf{x}_l = h_l(\mathbf{x}_{l-1}; \mathbf{w}_l),
\end{align}
where $\mathbf{x}_{l-1}$ is the input of the $l$th layer
and also the output of the $(l-1)$th layer, or is
the input of the whole network if $l=1$.
The representation extraction part of the network
can be written in a function form,
$H_L(\mathbf{x}_{0}) = h_L(H_{L-1}(\mathbf{x}_{0}))$,
where $H_l(\mathbf{x}_{0}) = h_l(H_{l-1}(\mathbf{x}_{0}))$.

\subsection{Shallow Fusion}
Network fusion is a process of combining multiple base networks,
e.g., $K$ base networks $\{H^1_{L_1}, \dots, H^K_{L_K}\}$\footnote{For simplicity, we only use the representation extraction part $H_{L_k}^k$ to describe a full network.}.
The conventional fusion in general includes two approaches:
feature fusion, fusing the representations extracted from the networks together,
followed by a classification layer;
and decision fusion (a.k.a., model ensemble),
fusing the classification scores computed from the networks.
This paper focuses on feature fusion and the fusion
can be formulated in the function form,
$
H(\mathbf{x}_{0}) = F(H^1_{L_1}(\mathbf{x}_{0}), \dots, H^K_{L_K}(\mathbf{x}_{0}))
$,
where the fusion function $F(\cdot)$,
in this paper,
is the sum of the representations,
\begin{align}
F(H^1_{L_1}(\mathbf{x}_{0}), \dots, H^K_{L_K}(\mathbf{x}_{0})) = \sum\nolimits_{k=1}^KH^k_{L_k}(\mathbf{x}_{0}).
\label{eqn:summationfusion}
\end{align}
The fusion function could be in other forms,
e.g., concatenation or maximization,
and we will discuss them later.
%\begin{align}
%F(H^1_{L_1}(\mathbf{x}_{0}), \dots, H^K_{L_K}(\mathbf{x}_{0})) = [(H^1_{L_1}(\mathbf{x}_{0}))^\top, \dots, (H^K_{L_K}(\mathbf{x}_{0}))^\top]^\top,
%\end{align}

\subsection{Deep Fusion}
Deep fusion performs feature fusion not only
over the final feature representation
but also over the intermediate feature representations.
The forward process and backward propagation are presented as follows.

\noindent\textbf{Forward process.}
A network $H^k_L$ with $L$ feature extraction layers is divided into $B$ blocks,
$\{G^k_1, \dots, G^k_b, \dots, G^k_B\}$
where a block $G^k_b$ is a sequence of several nonlinear transformations,
$\{h^k_{b_{start}}, \dots, h^k_{b_{end}}\}$:
$$G^k_b=h^k_{b_{end}}(h^k_{b_{end-1}}(\dots(h^k_{b_{start}}(\mathbf{x}_{b-1}))\dots)),$$
or simply an identity connection.
Deep fusion
is a process
of fusing $K$ networks
(assuming each consisting of $B$ blocks),
with $B$ summation fusion operations:
\begin{align}
\bar{\mathbf{x}}_1 ~&=~ F_1(G^1_1(\mathbf{x}_0), G^2_1(\mathbf{x}_0), \dots, G^K_1(\mathbf{x}_0)), \\
\bar{\mathbf{x}}_2 ~&=~ F_2(G^1_2(\bar{\mathbf{x}}_1), G^2_2(\bar{\mathbf{x}}_1), \dots, G^K_2(\bar{\mathbf{x}}_1)),\label{eqn:deepsumfusion:2} \\
\dots \\
\bar{\mathbf{x}}_B ~&=~ F_B(G^1_B(\bar{\mathbf{x}}_{B-1}), G^2_B(\bar{\mathbf{x}}_{B-1}), \dots, G^K_B(\bar{\mathbf{x}}_{B-1})),
\label{eqn:deepsumfusion}
\end{align}
where $F_b(G^1_b(\bar{\mathbf{x}}_{b-1}), G^2_b(\bar{\mathbf{x}}_{b-1}), \dots, G^K_b(\bar{\mathbf{x}}_{b-1}))
= \sum_{k=1}^K G^k_b(\bar{\mathbf{x}}_{b-1})$,
$\bar{\mathbf{x}}_b$ is the output of the $b$th fusion,
and it is assumed that
the output sizes of the $K$ blocks $G^k_b(\bar{\mathbf{x}}_{b-1})$
are the same.
Figure~\ref{fig:compareshallowanddeep}
illustrates the difference between shallow and deep fusions.

\noindent\textbf{Backward propagation.}
Gradient back-propagation is
the same as the conventional back-propagation.
Here, we present the back-propagation form
with respect to the input and the output of each fusion,
i.e., over the blocks.
According to the definition,
the gradient of $\bar{\mathbf{x}}_{b+1}$
with respect to $\bar{\mathbf{x}}_b$,
called fused block gradient,
can be computed as follows,
\begin{align}
\frac{\partial \bar{\mathbf{x}}_{b+1}}{\partial \bar{\mathbf{x}}_b}
= \sum\nolimits_{k=1}^K \frac{\partial G^k_{b+1}}{\partial \bar{\mathbf{x}}_b},
\end{align}
which intuitively means that the gradient is
the summation
of block gradients $\{\frac{\partial G^k_{b+1}}{\partial \bar{\mathbf{x}}_b}\}_{k=1}^K$,
and block gradient $\frac{\partial G^k_{b+1}}{\partial \bar{\mathbf{x}}_b}$
is the gradient of the $(b+1)$th block of the $k$th base network.
Suppose the loss function is $O(\bar{\mathbf{y}}, \mathbf{y}^*)$,
with $\bar{\mathbf{y}}$ and $\mathbf{y}^*$
being the estimated and ground-truth labels.
The gradient with respect to the hidden response $\bar{\mathbf{x}}_b$
using back-propagation
can be computed in the following form,
\begin{align}
\frac{\partial O}{\partial \bar{\mathbf{x}}_b}
=\frac{\partial \bar{\mathbf{x}}_{b+1}}{\partial \bar{\mathbf{x}}_b} \frac{\partial O}{\partial \bar{\mathbf{x}}_{b+1}}
= \prod\nolimits_{t=b}^{B-1}\frac{\partial \bar{\mathbf{x}}_{t+1}}{\partial \bar{\mathbf{x}}_t} \frac{\partial O}{\partial \bar{\mathbf{x}}_{B}}
= \prod\nolimits_{t=b}^{B-1} (\sum_{k=1}^K \frac{\partial G^k_{t+1}}{\partial \bar{\mathbf{x}}_t}) \frac{\partial O}{\partial \bar{\mathbf{x}}_{B}}.
\label{eqn:gradientfordeepfusion}
\end{align}

%\begin{align}
%\frac{\partial O}{\partial \bar{\mathbf{x}}_b}
%&~=~ \frac{\partial \bar{\mathbf{x}}_{b+1}}{\partial \bar{\mathbf{x}}_b} \frac{\partial O}{\partial \bar{\mathbf{x}}_{b+1}} \label{eqn:gradientfordeepfusion:2}\\
%&~=~ \prod_{t=b}^{B-1}\frac{\partial \bar{\mathbf{x}}_{t+1}}{\partial \bar{\mathbf{x}}_t} \frac{\partial O}{\partial \bar{\mathbf{x}}_{B}} \nonumber\\
%&~=~ \prod_{t=b}^{B-1} (\sum_{k=1}^K \frac{\partial G^k_{t+1}}{\partial \bar{\mathbf{x}}_t}) \frac{\partial O}{\partial \bar{\mathbf{x}}_{B}}.
%\label{eqn:gradientfordeepfusion}
%\end{align}
\noindent\textbf{Base network selection.}
From the above descriptions,
we can see that
the computation complexity for both the forward and backward processes
is almost equal to the complexity of all the base networks
with the negligible element-wise addition cost.
Therefore,
deep fusion does not introduce additional parameters, nor increase the computation complexity.

Deep fusion typically chooses a very deep network, and a shallower (might also be deep) network
in which each block contains only a convolution layer,
with a few blocks/fusions
(for example, each block corresponds to one scale).
Consequently, the computation complexity
is approximately equal to that of the very deep network.
This nice property makes deep fusion comparable
to the plain network (i.e., the deep one used for deep fusion),
Highway~\cite{SrivastavaGS15a}, and ResNet~\cite{HeZRS15b}
that includes some non-identity connections whose extra cost is similar to ours.
In addition,
training the fused net formed from a very deep base network and a shallow base network
is less difficult than training the initial very deep base network
because
the fused net could
be composed from
a less deep base network,
and a less shallow base network,
which will be shown later,
and
the essential depth of a fused net
is reduced.

Such a choice resembles the teacher and student framework (e.g., FitNets~\cite{RomeroBKCGB14} and Net2Net~\cite{ChenGS15})
where the student network is learnt from the already-trained teacher network.
But in our approach,
the teacher (shallow) and student (deep) networks are jointly learnt
and benefit from each other.
And the vanishing gradient problem if it seriously exists for the deep base network,
is alleviated, according to the gradient back propagation shown in Equation~(\ref{eqn:gradientfordeepfusion}).
Besides,
we will show that such a choice
is advantageous in the flow of information.

\subsection{Analysis}
\begin{figure}[t]
\centering
(a)~{\includegraphics[width=.45\linewidth]{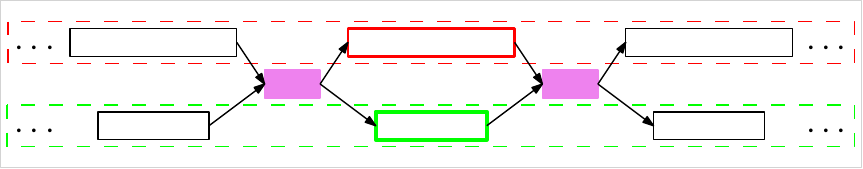}}
(a)~{\includegraphics[width=.45\linewidth]{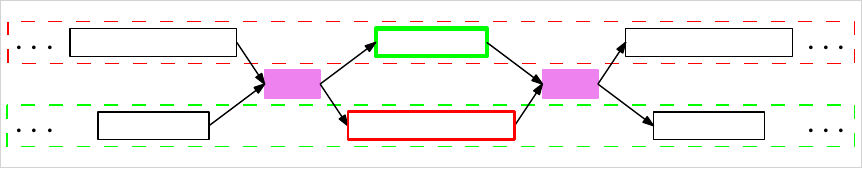}}
\caption{Illustrating that the same fused net can
be composed of a group of $2$ base networks shown in (a)
and another group of $2$ different base networks shown in (b).
The difference lies in that the middle block is exchanged.
Similarly, the third block can be exchanged.
Thus, many different base network groups can form the same fused net. }
\label{fig:deepfusionequivalence}
\vspace{-.3cm}
\end{figure}

\noindent\textbf{High capability
of combining multi-scale representations.}
A deeply-fused net composed from a group of $K$ base networks
can also be composed from another group of $K$ different base networks,
which is shown below.
Considering the mathematical formulation of deep fusion,
we can rewrite Equation~(\ref{eqn:deepsumfusion:2})
into an equivalent form,
\begin{align}
\bar{\mathbf{x}}_2 = F_2({\color{red}G^1_2}(\bar{\mathbf{x}}_1), {\color{green}G^2_2}(\bar{\mathbf{x}}_1), \dots, G^K_2(\bar{\mathbf{x}}_1))
=F_2({\color{green}G^2_2}(\bar{\mathbf{x}}_1), {\color{red}G^1_2}(\bar{\mathbf{x}}_1),  \dots, G^K_2(\bar{\mathbf{x}}_1)), \nonumber
\end{align}
which does not change the fused net.
This property is called block exchangeability.
It can be regarded as
changing the first two base networks:
$G_1^1 \rightarrow {\color{red}G_2^1} \rightarrow \dots \rightarrow G_B^1$
and $G_1^2 \rightarrow {\color{green}G_2^2} \rightarrow \dots \rightarrow G_B^2$,
to two other base networks:
$G_1^1 \rightarrow {\color{green}G_2^2} \rightarrow \dots \rightarrow G_B^1$
and $G_1^2 \rightarrow {\color{red}G_2^1} \rightarrow \dots \rightarrow G_B^2$.
Similarly, we can obtain more base networks, but resulting in the same fused net.
The number of unique combinations of $K$ base networks can reach up
to $(K!)^{B-1}$(In practice the number will be smaller than $(K!)^{B-1}$, but it is still very large).
Figure~\ref{fig:deepfusionequivalence} shows an example
to illustrate this block-exchangeable property.

Each possible base network, in our implementation,
will output an $8 \times 8$ feature map,
in which each element corresponds to a receptive field
with the size (scale) depending on the base network.
The fused net can be formed
from many base networks,
and thus there exist many receptive fields
with various sizes.
In addition,
the two base networks, with same sizes of receptive field,
may have different extraction processes,
and thus the extracted representations
are different, and are able to capture different characteristics.

\noindent\textbf{Improvement of the information flow.}
We show that an earlier intermediate layer might have a shorter path
to the output layer.
Consider an early intermediate representation, e.g., $\bar{\mathbf{x}}_1$,
the shortest path to the final feature representation
is $\sum_{b=2}^{B} \min_{k=1, \dots, K} |G_b^k|$,
which intuitively means that for each fused block
the smallest block is chosen as the path.
This implies that
the path from the intermediate layer in the deeper base network to the output becomes shorter,
%in our suggested choice: one shallow base network
%and one deep base network,
and thus the supervision
can be fast flowed to an early intermediate layer.

Similarly,
a later intermediate layer may have a shorter path
from the input layer,
indicating that
the input information can be quickly fed into a later intermediate layer
instead of
through a long path.
This benefit to the network learning
is in some sense related to relay back-propagation~\cite{ShenLH15},
which explicitly fixes the earlier layers
(something like directly connecting the input
to the later layers)
when updating the later layers.
In summary,
the flow of information
from the input to the intermediate layers
and from the intermediate layers to the output
are both improved,
which is beneficial to training a deep network.

\section{Discussions}
\noindent\textbf{Concatenation, Maximization, and Summation.}
With the summation fusion in the intermediate layers,
there is almost no change for each base network: the network structures are not changed.
The only effect is that the output is changed with some signals added
from other networks.
Maximization fusion
that performs an element-wise maximization,
$$F(H^1_{L_1}(\mathbf{x}_{0}), \dots, H^K_{L_K}(\mathbf{x}_{0})). = \max\limits_{k=1, \dots, K} \{H^k_{L_k}(\mathbf{x}_{0}).\}, $$
is studied in~\cite{LiaoC15a}.
Similar to summation fusion,
there is almost no change for each base network.
In contrast,
with the concatenation fusion,
the fusion function Equation~(\ref{eqn:summationfusion}) is
changed to
$$F(H^1_{L_1}(\mathbf{x}_{0}), \dots, H^K_{L_K}(\mathbf{x}_{0})) = [(H^1_{L_1}(\mathbf{x}_{0}))^\top, \dots, (H^K_{L_K}(\mathbf{x}_{0}))^\top]^\top,$$
e.g., an inception module in GoogLeNet.
The base networks have to be changed
and more parameters are needed:
the input size of the subsequent sub-network immediately after the fusion in each base network is increased
as the fusion output becomes larger
(or in the original base network,
there are many channels with all $0$ entries appended
in the output of a block
so that the total output matches the size with the input of the subsequent inception module).
The combination of concatenation fusions and summation fusions
is studied in~\cite{SzegedyIV16}, which shows better ImageNet performances.
%Figure~(\ref{fig:comparesummationandconcatenation})
%illustrates the difference.

%INCEPTION-V4, INCEPTION-RESNET AND THE IMPACT OF RESIDUAL CONNECTIONS ON LEARNING.
%ChristianSzegedy&SergeyIoffe&VincentVanhoucke

%\begin{figure}[t]
%\centering
%\subfigure[Sum/max fusion]{\includegraphics[]{figs/SummationFusion}}~~~
%\subfigure[Concatenation fusion]{\includegraphics[]{figs/ConcatenationFusion}}
%\caption{The output of summation (maximization) fusion can be directly fed into the original subsequent sub-network
%without changing the network,
%while the output of concatenation fusion
%cannot be directly fed into the original subsequent sub-network
%as it is larger,
%and needs more parameters.}
%\label{fig:comparesummationandconcatenation}
%\end{figure}

\begin{figure}[t]
\centering
\subfigure[A deeply-supervised net]{\includegraphics[scale=0.8]{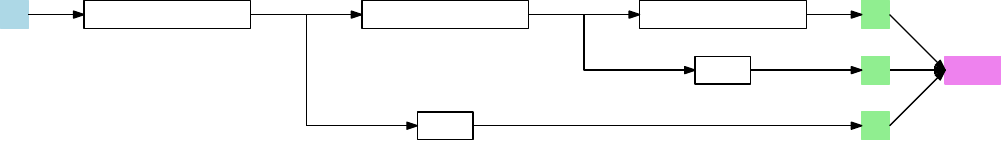}}\\
\subfigure[A network equivalent to the deeply-supervised net]{\includegraphics[scale=0.8]{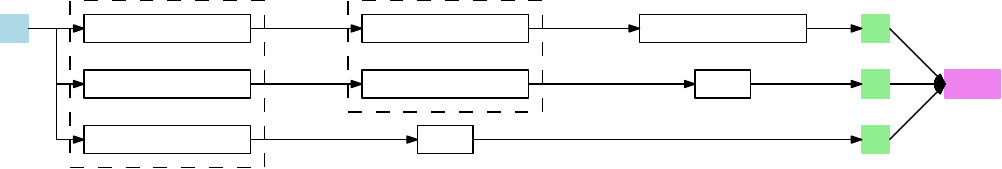}}\\
\subfigure[Unidirectional fusion]{\includegraphics[scale=0.8]{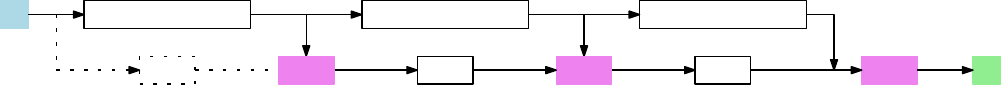}}\\
\caption{
(a) a deeply-supervised net;
(b) the shallow fusion view of the deeply-supervised net:
the blocks in the dashed box share the same structure and parameters;
(c) Unidirectional deep fusion can be regarded
as an alternative of a deeply-supervised net.
The dotted part could be removed
if only the intermediate outputs,
without the input, are flowed
from the main network $H_o$
to the below network $H_i$.}
\label{fig:compareudfanddsn}
\end{figure}

\noindent\textbf{Relation to Deeply-Supervised Nets.}
The deeply supervised net estimates the network parameters
through optimizing multiple losses,
some of which come from the intermediate layers.
The formulation could be written as follows,
\begin{align}
\ell(C(H(\mathbf{x}_0), \mathbf{y}^*) + \sum\nolimits_{b=1}^B \ell (C_b(H'_{b}(\mathbf{x}_0)), \mathbf{y}^*),
\end{align}
where $H'_{b}(\mathbf{x}_0)$ is a subnetwork of $H$
and consists of the part from the input to the layer $h_{b_{end}}$,
and $C(\cdot)$ and $C_b(\cdot)$ are the classifiers.
A similar network used for edge detection~\cite{XieT15}
is shown to be able to combine multi-scale information.
This formulation can be interpreted
as a shallow decision/loss fusion process:
combine $(B+1)$ networks
with shared parameters across the $(B+1)$ networks
$\{H'_1, \dots, H'_B, H\}$,
which is illustrated in Figure~\ref{fig:compareudfanddsn}(b).

In addition,
we show that the unidirectional version of deeply fused network
is closely related to
the deeply-supervised net with weights sharing for the classification layers.
The unidirectional deeply fused network,
(combining two networks,
the signal from the network $H_o$ is flowed to the network $H_i$) is mathematically
given as follows,
\begin{align}
\mathbf{x}^o_1 &~=~ G^o_1(\mathbf{x}_0),~ \mathbf{x}^i_1 ~=~ \mathbf{x}^o_1 + G^i_1(\mathbf{x}_0), \\
\mathbf{x}^o_2 &~=~ G^o_2(\mathbf{x}^o_1),~ \mathbf{x}^i_2 ~=~ \mathbf{x}^o_2 + G^i_1(\mathbf{x}^i_1), \\
\dots \\
\mathbf{x}^o_B &~=~ G^o_B(\mathbf{x}^o_{B-1}),~ \mathbf{x}^i_B ~=~ \mathbf{x}^o_B + G^i_1(\mathbf{x}^i_{B-1}),
\end{align}
and a classification layer, $C^i(\mathbf{x}^i_B)$, is defined over $\mathbf{x}^i_B$.

Figure~\ref{fig:compareudfanddsn}(c)
shows an example of
unidirectional deep fusion.
Compared with the deeply-supervised net in Figure~\ref{fig:compareudfanddsn}(a),
we can observe that the unidirectional deep fusion uses progressive feature fusion,
while deep supervision in deeply-supervised nets uses loss fusion.
% {\color{red}Besides, the feature fusion in our approach is progressively conducted:
%the fused feature of an early sub-network $H_e$ is fused together with the output of the immediately-subsequent block $b$
%yielding a fused feature of the new early sub-network with the new block included ($H_e + b$).}

%\begin{figure}
%\centering
%\includegraphics[]{figs/SkipLayerConnection}
%\caption{An example interpreting the skip-layer connection as a kind of unidirectional deep fusion.}
%\label{fig:skiplayerconnection}
%\end{figure}

\noindent\textbf{Relation with Highway and ResNet.}
Skip-layer connection
means that
a layer can take not only the layer at the previous level as input but also some of the lower layers.
It resembles deep fusion and in some sense
they can be equivalently transformed to each other.
Here we consider two recently-developed network
examples: Highway~\cite{SrivastavaGS15a} and ResNet~\cite{HeZRS15b}.

\begin{figure}[t]
(a)~\includegraphics[scale=0.8]{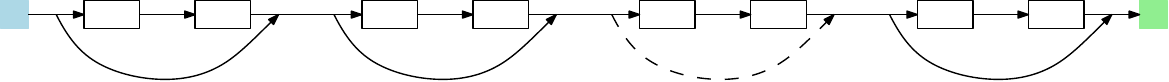}\\
(b)~\includegraphics[scale=0.8]{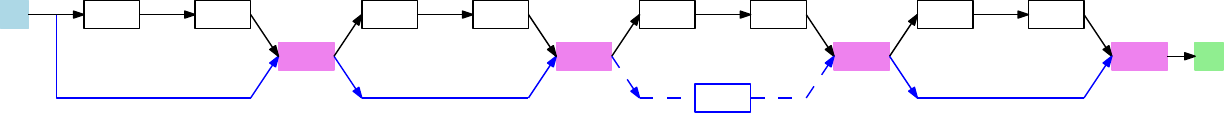}
\caption{Illustrating that ResNet is an example of deep fusion.
The solid skip-layer connection is an identity connection,
and the dashed skip-layer connection is a linear projection.
From (b), we can see that
ResNet is a fused network from the plain network
and a short network (highlighted in blue).}
\label{fig:resnet}
  %\vspace{-.1cm}
\end{figure}

Highway and ResNet can be viewed as combining two networks:
one is deep, called a plain network,
and the other one is shallow,
a sequence of virtual layers (identity connection)
and possible extra layers,
which are just down-sample pool layers (same to the plain network) in Highway,
and are projection layers in ResNet
(e.g., the blue box means a linear projection in Figure~\ref{fig:resnet}).
Figure~\ref{fig:resnet} illustrates that ResNet
is an example of a deeply-fused net when the number of base networks is $2$.
Similarly, Highway can be transformed to deep fusion,
and the difference is that the fusion in Highway is a weighted sum
and the weight is data customized through the transform gate.

The block in the deep/plain network in Highway and ResNet is typically small,
consisting of $1$-$3$ layers.
And thus there are many blocks/fusions,
while in our approach the number of fusions is suggested smaller.
We use the non-identity connection
to form the block in the shallow network:
usually one block in one scale (could be more in a very deep network), and
no extra layer except pooling layer
across scales;
while ResNet uses non-identity connection
to match the size which is changed when across scales in the network.
Thus the number of parameters in ResNet and our approach are similar
and both are smaller than that in Highway.

\begin{table}[t]
\centering
\scriptsize
\caption{Base network architectures.
In our implementation,
the blocks are formed
with the layers before FC1
(but in general, the block could also include FC1),
and FC1 and Ip1 together are called the classification layer in this paper.}
\label{tab:networkarchitectureBN}
  \begin{tabular}{c|c|c||c|c||c|c|c|c|c}
  \hline
  \multicolumn{3}{c||}{ Network } & N1  & N2 &  N3  & N4  & N5 & N6 & N7 \\
  \hline
 \multicolumn{3}{c||}{ \#Layers} & $19$ & $50$ & $5$ & $8$ & $10$ & $11$ & $14$  \\
  \hline
  \hline
  ~Layer name~ & ~Output size~ & Parameters & \multicolumn{7}{c}{Repeat times}\\
  \hline
C$1.$  & $32 \times 32$ &$(3\times 3, 32)$ & $\times5$ &  $\times16$ & $\times1$ & $\times2$ & $\times2$ & $\times3$ & $\times4$  \\
  \hline
 \multicolumn{10}{c}{ $2\times 2$ max pool, stride $2$} 　 \\
  \hline
 C$2.$ & $16 \times 16$ & $(3\times 3,80) $ & $\times6$ &  $\times16$ & $\times1$ & $\times2$ & $\times3$ & $\times3$ & $\times4$ \\
  \hline
 \multicolumn{10}{c}{ $2\times 2$ max pool, stride $2$} 　 \\
  \hline
  C$3.$ & $8 \times 8$ & $(3\times 3,128)$ & $\times6$ &  $\times16$ & $\times1$ & $\times2$ & $\times3$ & $\times3$ & $\times4$  \\
  \hline
  FC$1$ & $8 \times 8$ & $(1\times 1,100)$ & $\times1$ & $\times1$ & $\times1$ & $\times1$ & $\times1$ & $\times1$ &  $\times1$\\
\hline
   \multicolumn{10}{c}{ $8\times 8$ avg pool, stride $8$} 　 \\
    \hline
 Ip$1$ & $1 \times 1$ & $(1\times 1,100$ or $10)$ & $\times1$ & $\times1$ & $\times1$ & $\times1$ & $\times1$ & $\times1$ &  $\times1$ \\
  \hline
  \end{tabular}
  \vspace{-.3cm}
\end{table}

\section{Experiments}
We evaluate our approach
on the CIFAR-10 and CIFAR-100 datasets.
The CIFAR-10 and CIFAR-100 datasets~\cite{KrizhevskyH09}
are both subsets drawn from the $80$-million tiny image database~\cite{TorralbaFF08}.
The CIFAR-10 dataset consists of $60000$ $32\times 32$ colour images in $10$ classes, with $6000$ images per class.
There are $50000$ training images and $10000$ test images.
The CIFAR-100 dataset is like the CIFAR-10,
except that it has $100$ classes each containing $600$ images.
There are $500$ training images and $100$ testing images per class.
The $100$ classes in the CIFAR-100 are grouped into $20$ superclasses.
Each image comes with a "fine" label (the class to which it belongs)
and a "coarse" label (the superclass to which it belongs).
%In both datasets, training and testing images are uniformly distributed over all the categories.

The architecture of the base network is built upon
basic units: the convolution layer,
the nonlinear ReLU activation function,
the max-pooling layer,
the fully-connected layer,
and the softmax layer for training,
which are intentionally chosen
to directly show the benefits from deep fusion.
We also use a batch normalization layer right after each convolution layer.
The details of the network architectures used in our experiments
are presented in Table~\ref{tab:networkarchitectureBN} (base networks)
and Table~\ref{tab:networkarchitectureDFN} (block division).

We train the networks using the SGD algorithm,
with the weight decay regularization,
the momentum set to $0.9$,
the mini-batch size set to $100$,
and the maximum number of epochs set to $400$.
An exponentially decay learning rate is used:
the learning rate is reduced by a factor of $10$ after the $200$th, $300$th, $350$th epoch.
The results of our approach, the baseline algorithms and the plain network
are reported
with the weight decay coefficient and the initial learning rate
tuned in the ranges:
$\{0.0001, 0.0005\}$ and $\{0.01, 0.1\}$ for $19$ layers,
and $\{0.0002, 0.0005, 0.001\}$ and $\{0.02, 0.05, 0.1\}$ for $50$ layers.
The weights are initialized using the scheme~\cite{HeZRS15a}.
Experiments are conducted using Caffe~\cite{Jia_2014_CAFFE}.
The datasets are preprocessed using a common setting, as described in~\cite{SrivastavaGS15a},
including the global contrast normalization,
four-zero pixels padding at all borders with $32\times 32$ random crops,
and random horizontal flipping.

\begin{table}[t]
\centering
\scriptsize
\caption{Block division.
If one block starts from C21 or C31,
the block also includes the preceding pooling layer
(in implementation, we only perform max pooling one time for all the base networks
as the input and the operation are the same),
and for clarity this is not explicitly described in the table.
N33, N46, and N58 are shallow networks,
with each block containing only one convolution layer.
We use N13N33 to represent
a fused net formed
from base networks N1 and N3
with block division N13 and N33.}
\label{tab:networkarchitectureDFN}
   \begin{tabular}{c|c|c|c|c||c|c|c||c|c|c}
   \hline
  N13   & N33 &  N43 & N63 &  N73 & N16 & N26 & N46  & N18  &  N28 &  N58 \\
     \hline
     \hline
  \multicolumn{5}{c||}{$3$ blocks} & \multicolumn{3}{c||}{$6$ blocks} & \multicolumn{2}{c}{$8$ blocks}\\
 \hline
  \multirow{3}{*}{C11-C15} &
  \multirow{3}{*}{C11} &
    \multirow{3}{*}{C11-C12} &
    \multirow{3}{*}{C11-C13} &
      \multirow{3}{*}{C11-C14} &
   \multirow{2}{*}{ C11-C12} &
    \multirow{2}{*}{C11-C18} &
     \multirow{2}{*}{C11} &
   \multirow{2}{*}{ C11-C12}  &
   \multirow{2}{*}{C11-C18} &
   \multirow{2}{*}{C11} \\
         &  &  & &   & &  & & & &  \\
         \cline{6-11}
       & & & &  & C13-C15 & C19-C116 & C12  & C13-C15
       & C19-C116 & C12      \\

       \hline
  \multirow{3}{*}{C21-C26} &
  \multirow{3}{*}{C21} &
       \multirow{3}{*}{C21-C22} &
       \multirow{3}{*}{C21-C23} &
            \multirow{3}{*}{C21-C24}&
      \multirow{2}{*}{ C21-C23} &
      \multirow{2}{*}{ C21-C28} &
           \multirow{2}{*}{C21} &
    C21-C22  &
    C21-C25 &
      C21 \\
      \cline{9-11}
            & & & & & & & & C23-C24  & C26-C210 & C22   \\
            \cline{6-11}
          & & & & & C24-C26 & C29-C216 & C22  & C25-C26
         & C211-C216 & C23     \\

          \hline
           \multirow{3}{*}{C31-C36} &
             \multirow{3}{*}{C31} &
                  \multirow{3}{*}{C31-C32} &
                  \multirow{3}{*}{C31-C33} &
                       \multirow{3}{*}{C31-C34} &
      \multirow{2}{*}{ C31-C33} &
      \multirow{2}{*}{C31-C38} &
           \multirow{2}{*}{C31} &
    C31-C32  &
     C31-C35 & C31 \\
      \cline{9-11}
            &  & &  & & & & & C33-C34  & C36-C310 & C32    \\
            \cline{6-11}
         & & & &  & C34-C36  & C39-C316 & C32  & C311-C316 & C35-C36
        & C33      \\
            \hline
   \end{tabular}
    \vspace{-.5cm}
\end{table}

\subsection{Empirical Analysis}
\begin{figure*}[t]
\centering
(a)~{\includegraphics[width=.4\linewidth]{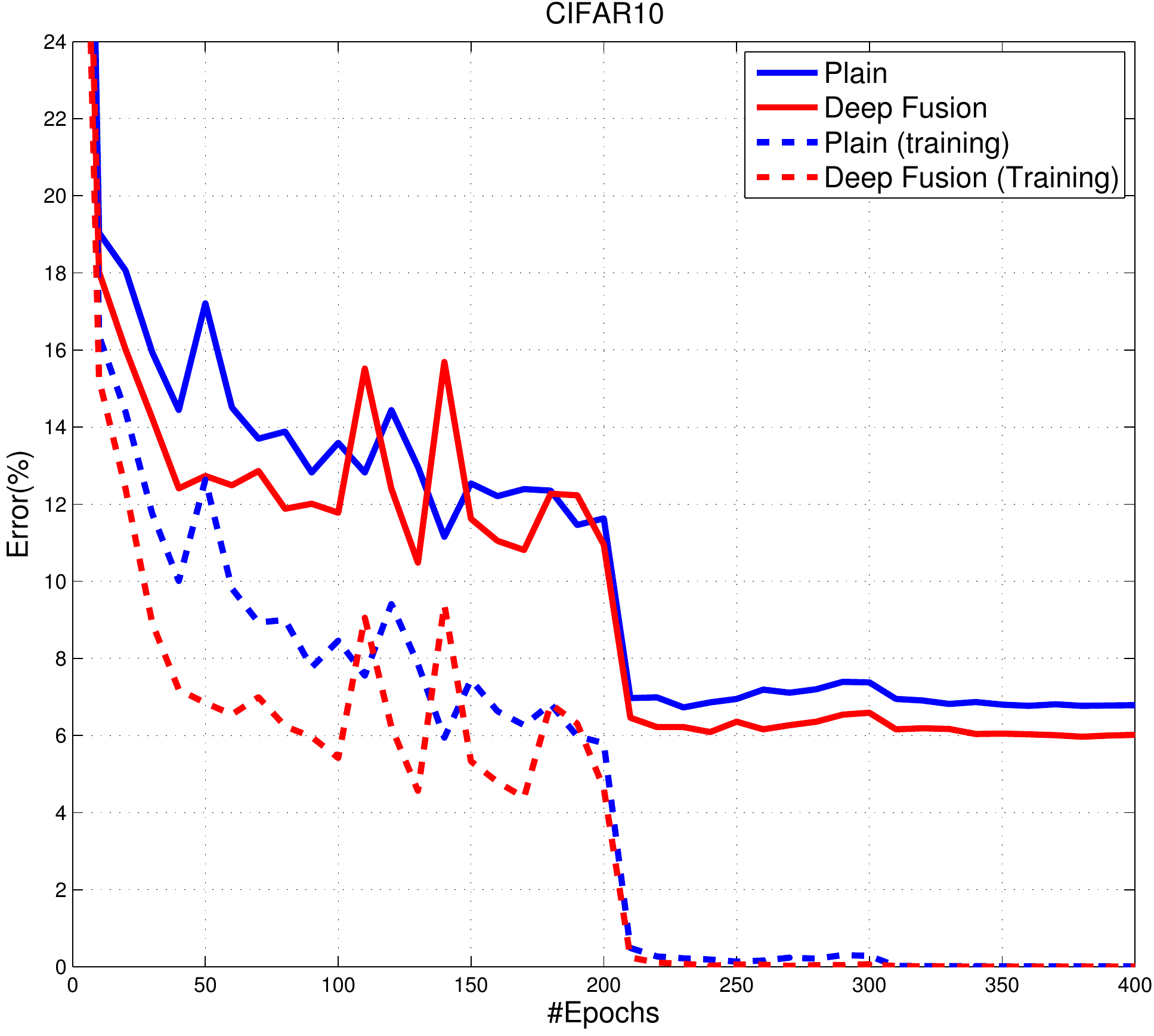}}~~~
(b)~{\includegraphics[width=.4\linewidth]{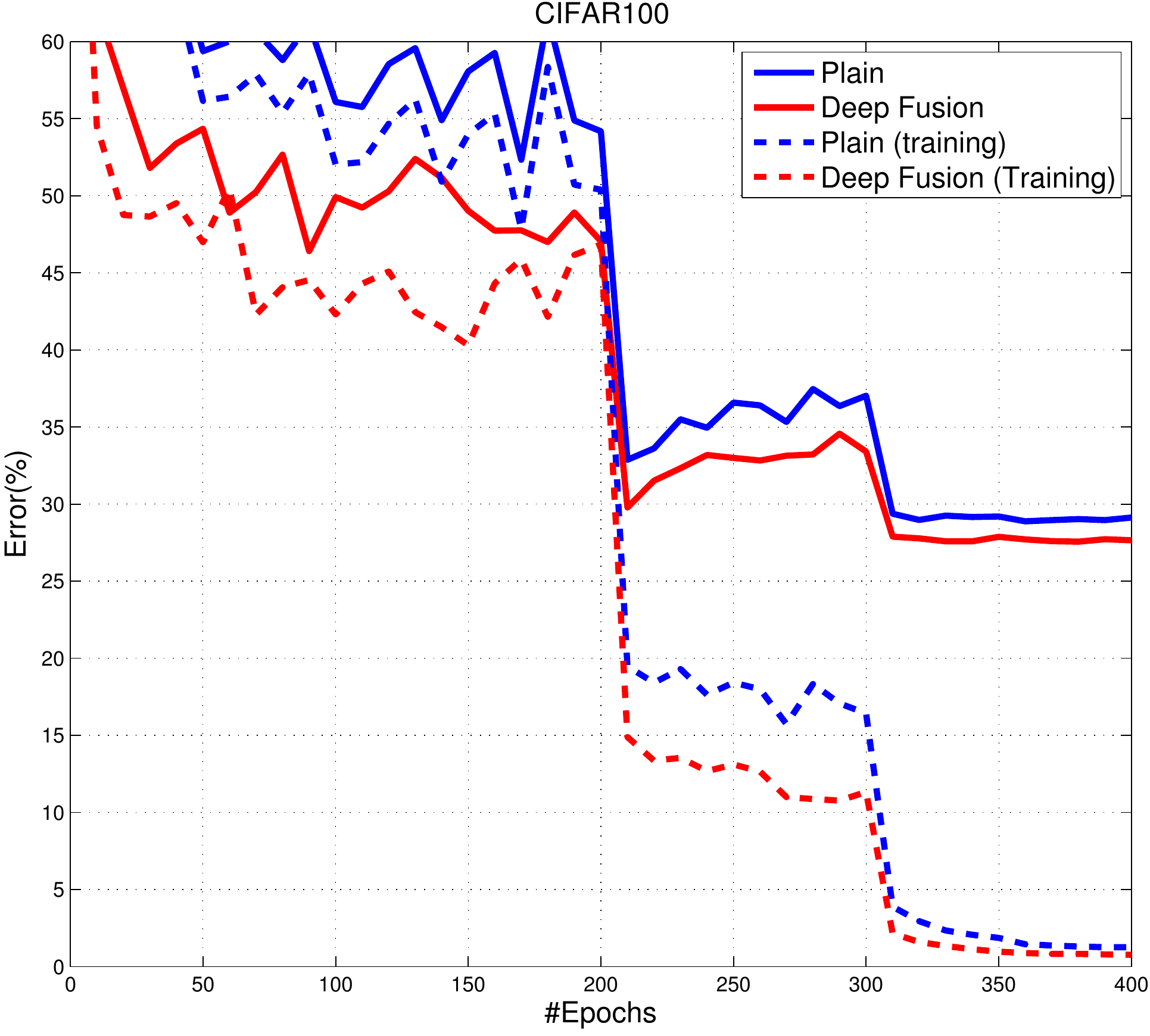}}
\caption{Training and testing errors ($\%$) vs. \#epoches for the plain network
and the deeply-fused net (N13N33) on  (a) CIFAR-10 and (b) CIFAR-100.}
\label{fig:convergencecurve}
  \vspace{-.2cm}
\end{figure*}

\noindent\textbf{Convergence curve.}
Figure~\ref{fig:convergencecurve} shows the training error and the testing error
between a fused net, N13N33,
composed of two base networks N13 and N33 (See Table~\ref{tab:networkarchitectureDFN}),
and a plain network, N13, the deeper base network in the fused net.
It can be observed that both the training error and testing error
of deep fusion are lower than the plain network in the later iterations,
and deep fusion achieves the same error with plain network using much fewer training steps.

\begin{table}[t]
\centering
\scriptsize
\caption{Accuracy ($\%$) comparison between different fusion methods.}
\label{tab:deepvsshallow}
\begin{tabular}{c|c|c}
\hline
& ~CIFAR-10~ & ~CIFAR-100~  \\
\hline
\multicolumn{3}{c}{$19$ layers (N13N33) } \\
\hline
Plain & $93.5$ & $70.87$ \\
Deep concatenation fusion& $93.4$ & $70.64$  \\
Deep max fusion& $93.14$ & $69.55$  \\
Shallow fusion & $93.37$ & $71.23$  \\
Decision fusion (separate train)  & $93.48$ & $72.28$  \\
Decision fusion(joint train) & $93.06$ &  $71.11$\\
\hline
Deep summation (fusion before ReLU)  & $\mathbf{93.98}$ & $\mathbf{72.29}$  \\
Deep summation (fusion after ReLU) & $\mathbf{93.77}$ & $\mathbf{72.64}$  \\
\hline
\multicolumn{3}{c}{$50$ layers (N26N46)}  \\
\hline
Plain & $92.08$ & $65.48$ \\
Decision fusion (separate train)  & $93.39$ &  $68.9$\\
\hline
Deep summation   & $93.6$ & $72.39$   \\
\hline
\end{tabular}
  \vspace{-.5cm}
\end{table}

\noindent\textbf{Fusion.}
We present the results
from our deep summation fusion, deep max fusion,
deep concatenation fusion (the channel size of the convolution layer right before the pooling layer is reduced by half in order to match the size of the subsequent network),
shallow feature fusion (the feature fusion is only performed
at the final output of C3.),
decision fusion with the networks jointly trained
and with each network separately trained,
where the base networks are N13 and N33 for $19$ layers,
and are N26 and N46 for $50$ layers.
The comparison is given in Table~\ref{tab:deepvsshallow}.
It can be seen that
our approach achieves superior results.
There is a slight performance difference
when the fusion is conducted before or after ReLU,
and our other experiments are reported with fusion before ReLU.
It is interesting that the performance of decision fusion with separate training for $19$ layers on CIFAR-100
is also good,
but the performance on CIFAR-100 for $50$ layers is dramatically deteriorated.
In contrast,
our approach performs similarly for $19$ and $50$ layers,
which shows that our approach indeed can help train a very deep network.

\noindent\textbf{Performance with different \#blocks.}
We empirically study the performance with different number of blocks.
We consider three fused nets, N13N33, N16N46, N18N58,
in which the base networks have
$3$, $6$, and $8$ blocks respectively,
and the deep base network is $19$ layers
(see Table~\ref{tab:networkarchitectureDFN} for more details).
The comparison given in Table~\ref{tab:performancevsblocks}
shows that
the performances with different \#blocks
are close,
and over the challenging dataset CIFAR-100,
more blocks result in worse performance.
Thus, we will use $3$ blocks to compare with the baseline networks
as the computation complexity is almost the same as the plain network.

\begin{table}[t]
\centering
\scriptsize
\caption{Accuracy ($\%$) comparison with different \#blocks for deep fusion.}
\label{tab:performancevsblocks}
\begin{tabular}{c|c|c}
\hline
\#Blocks& ~CIFAR-10~ & ~CIFAR-100~  \\
\hline
3 (N13N33)  & $93.98$ & $72.64$  \\
6 (N16N46)&  $94.08$ & $72.02$  \\
8 (N18N58)& $94.01$ & $71.92$   \\
\hline
\end{tabular}
  %\vspace{-.2cm}
\end{table}

\iffalse
\noindent\textbf{Performance with different \#(base networks).}
The fused network can be formed from
$2$ base networks,
$3$ base networks,
and more base networks.
Table~\ref{tab:performancevsbasenetworks}
shows the performance
of the fused networks with different numbers of base networks,
with each containing three blocks (i.e., three fusions):
{\color{red}N13N23, ????.}

\begin{table}
\centering
\scriptsize
\caption{Accuracy ($\%$) with different \#(base networks) for deep fusion}
\label{tab:performancevsbasenetworks}
\begin{tabular}{l|c|c}
\hline
\#(Base networks)& CIFAR-10 & CIFAR-100  \\
\hline
2 (N13N33) & $93.98$ & $72.64$  \\
3 (N13N33N43)& $94.13$ & $72.92$ \\
4 (N13N33N43N63)& $94.01$ & $72.75$ \\
5 (N13N33N43N63N73)\\
\hline
\end{tabular}
\end{table}
\fi

\subsection{Comparison to Baseline Algorithms}
We compare our approach with three baseline networks:
plain network,
Highway~\cite{SrivastavaGS15a} and ResNet~\cite{HeZRS15b}.
The three baseline networks use the same plain network N1 with $19$ layers,
and our approach uses the plain network
as the deeper base network together with another base network N33
to form the deeply fused net, N13N33.
The number of parameters as well as the computation cost of our approach and ResNet are almost the same,
as (1) ResNet includes $8$ residual connections for the $19$-layer network,
consisting of two non-identity connections to match the dimensions,
and (2) there are three blocks in the shallow network in our approach,
and the first one is small and thus computational negligible
and the other two are almost the same with the  two non-identity connections in ResNet.
Highway includes more parameters as it introduces transform gates and hence is computationally more expensive.
In addition to the results from our implementation of the plain network,
Highway (using batch normalization w/o dropout) and ResNet
(batch normalization used before ReLU),
we also report the results of Highway and ResNet from the original papers.

\begin{table}[t]
\centering
\scriptsize
\caption{Accuracy ($\%$) comparison on CIFAR-10.}
\label{tab:CIFAR10Results}
\begin{tabular}{c|c|c|c}
\hline
Network & ~\#Layers~ & ~\#Parameters~ & ~Accuracy~\\
\hline
\hline
Plain & $19$ & $\sim1.20M$& $93.50$ \\
\hline
Highway (our implementation with dropout) & $19$ & $\sim2.26M$ & $93.35$\\
Highway (our implementation without dropout) & $19$ & $\sim2.26M$ & $93.06$\\
Highway~\cite{SrivastavaGS15a} & $19$ & $\sim2.3M$ & $92.46$  \\
Highway~\cite{SrivastavaGS15a} & $32$ & $\sim1.25M$ & $91.20$ \\
\hline
Resnet (our implementation)   & $19$ &  $\sim1.21M$ & $93.87$\\
Resnet~\cite{HeZRS15b}   & $20$ & $0.27M$ & $91.25$ \\
Resnet~\cite{HeZRS15b}   & $32$ & $0.46M$ & $92.49$ \\
Resnet~\cite{HeZRS15b}   & $44$ & $0.66M$ & $92.83$\\
Resnet~\cite{HeZRS15b}   & $56$ & $0.85M$ &  $93.03$ \\
Resnet~\cite{HeZRS15b}   & $110$ & $1.7M$ & $93.57$ \\
\hline
FitNet~\cite{RomeroBKCGB14} & $19$ & $\sim 2.5M$ &  $91.61$ \\

\hline
DFN (N13N33) & $19$ & $\sim1.31M$  & $\mathbf{93.98}$\\
\hline
\hline
Plain & $50$ & $\sim3.36M$ & $92.08$\\
\hline
DFN (N26N46) & $50$ & $\sim3.7M$ & $93.6$\\
DFN (N28N58) & $50$ & $\sim3.9M$ & $\mathbf{93.76}$\\
\hline
\end{tabular}
 % \vspace{-.5cm}
\end{table}

The results on CIFAR-10 and CIFAR-100
are shown in Table~\ref{tab:CIFAR10Results}
and Table~\ref{tab:CIFAR100Results}.
Compared with Highway that uses the gate
to select only a part of the plain network
for prediction,
our approach uses a small network to help train the whole plain network,
resulting in better performance.
In comparison to ResNet that uses identity-connection blocks
over the layers with the same dimension
and non-identity blocks over the layers with different dimensions
(appear when only scale changes in our $19$-layer network),
leading to too many blocks,
our approach uses non-identity connection
to form the block which is not across scales,
and has fewer blocks/fusions,
which might be the reason for superior performance.
Compared to the second best method, ResNet,
our approach achieves more significant gain on CIFAR-100
than CIFAR-10.

In addition,
we also report the performance
with a much deeper base network,
a $50$-layer network
in Table~\ref{tab:CIFAR10Results} and Table~\ref{tab:CIFAR100Results}.
We have several observations.
On the one hand,
the performance from $50$ layers for the plain network
is lower than that from $19$ layers
(e.g., decreased to $65.48\%$ from $70.87\%$ on CIFAR-100 ),
while the performances of our approach from $50$ layers and $19$ layers
($72.56\%$ and $72.64\%$)
are only slightly different.
On the other hand,
on CIFAR-10, our approach with $50$ layers performs  better than
ResNet with similar depth ($44$ and $56$ layers),
but ResNet has less parameters,
showing that our approach is helpful for training a deep and complex network.
Considering that the $50$-layer network has much more parameters
than the $19$-layer network,
deep fusion indeed helps train very deep network even with more parameters.

\begin{table}[t]
\centering
\scriptsize
\caption{Accuracy ($\%$) comparison on CIFAR-100.}
\label{tab:CIFAR100Results}
\begin{tabular}{c|c|c|c}
\hline
Network & ~\#Layers~ & ~\#Parameters~ & ~Accuracy~\\
\hline
\hline
Plain & $19$ &  $\sim1.21M$ & $70.87$ \\
\hline
Highway (our implementation with dropout) & $19$ & $\sim2.27M$ & $68.98$\\
Highway (our implementation without dropout) & $19$ & $\sim2.27M$ & $67.97$\\
Highway~\cite{SrivastavaGS15a} & $19$ & $\sim2.3M$ & $67.76$ \\
\hline
Resnet (our implementation)   & $19$ &  $\sim1.22M$ & $71.17$ \\
\hline
FitNet~\cite{RomeroBKCGB14} & $19$ & $\sim 2.5M$ & $64.96$ \\
\hline
DFN (N13N33) & $19$ & $\sim1.32M$  & $\mathbf{72.64}$\\
\hline
\hline
Plain & $50$ & $\sim3.36M$ & $65.48$\\
\hline
DFN (N26N46) & $50$ & $\sim3.7M$ & ${72.39}$\\
DFN (N28N58) & $50$ & $\sim3.9M$ & $\mathbf{72.48}$\\
\hline
\end{tabular}
 \vspace{-.3cm}
\end{table}

\iffalse
\begin{table}
\centering
\begin{tabular}{c|cc|cc}
method & CIFAR-10 & Rank & CIFAR-100 & Rank \\
\hline
fractional max-pooling~\cite{Graham14a} & $96.53\%$ & 1 & $73.61\%$  & 3 \\
\hline
all convnet~\cite{SpringenbergDBR14} & $95.59\%$ & 2 & $66.29\%$ &  19 \\
\hline
	All you need is a good init~\cite{MishkinM15} & $94.16\%$ & 3 & $72.34\%$ & 6 \\
\hline
Lessons learned from manually classifying CIFAR-10\footnote{http://karpathy.github.io/2011/04/27/manually-classifying-cifar10/} & $94\%$ & 4 & & \\
\hline
Generalizing Pooling Functions~\cite{LeeGT15} & $93.95\%$ & 5 & $67.63\%$ & 16 \\
\hline
	Spatially-sparse~\cite{Graham14} & $93.72\%$ & 6 & $75.7\%$ & 2 \\
\hline
	Scalable Bayesian Optimization~\cite{SnoekRSKSSPPA15} & $93.63\%$ & 7 & $72.60\%$ & 4 \\
\hline
	Deep Residual Learning ~\cite{HeZRS15} & $93.57\%$ & 8 & & \\
\hline
Fast and Accurate Deep Network Learning~\cite{ClevertUH15} & $93.45\%$ & 9 & $75.72\%$ & 1 \\
\hline
Competitive Multi-scale Convolutio\cite{LiaoC15a} & $93.13\%$ & 12 & $72.44\%$ & 5 \\
\hline
Batch-normalized~\cite{ChangC15} & $93.25\%$ & 11 & $71.14\%$ & 7 \\
\hline
\end{tabular}
\end{table}
\fi

\subsection{Comparison to State-of-the-Art Methods}
The results of
the top-performing algorthms
on CIFAR-10 and CIFAR-100 are
shown in Table~\ref{tab:stateofthearts}.
Overall speaking,
our approach
performs very well under the common data augmentation.
There are two competitive algorithms (both published in ICLR 2016),
LSUV~\cite{MishkinM15} that focuses on initialization
and ELU~\cite{ClevertUH15}
that focuses on a new nonlinear activation layer,
achieving similar performance as our approach.
We also report the results of our approach with more base networks (N13N33N43, N13N33N43N63N73):
both show better performance on CIFAR-100,
while on CIFAR-10 N13N33N43N63N73 does not make improvement.

Compared to LSUV~\cite{MishkinM15},
our approach (N13N33 and N16N46) performs better on CIFAR-100,
but worse on CIFAR-10.
On CIFAR-10, our approach performs the second best:
slightly lower than LSUV~\cite{MishkinM15} using maxout,
but greater than LSUV~\cite{MishkinM15} (19 layers) without using the maxout layer.
Our approach uses simple basic layers to show the benefit of deep fusion,
and we believe that potentially our approach can benefit from other advanced layers, e.g., maxout.

Compared to ELU~\cite{ClevertUH15},
our approach performs better on CIFAR-10,
but worse on CIFAR-100.
On CIFAR-100,
our approach performs the second best:
lower than ELU~\cite{ClevertUH15}
whose result is very high with common data augmentation.
ELU~\cite{ClevertUH15} introduces an exponential linear unit,
which is complementary to our approach and can be combined with our approach.

It is worth noting that
our deeply-fused net outperforms even much larger networks
with extreme data augmentation
like fractional max-pooling~\cite{Graham14a}.

\begin{table}[t]
\centering
\scriptsize
\caption{Accuracy ($\%$) comparison to state-of-the-art algorithms.}
\label{tab:stateofthearts}
\begin{tabular}{c|c|c}
\hline
\multicolumn{3}{c}{Common data augmentation}\\
\hline
Algorithm & ~CIFAR-10~ & ~CIFAR-100~ \\
\hline
DFN (N13N33) & $93.98$ & $\mathbf{72.64}$\\
DFN (N16N46) & $\mathbf{94.08}$  & $72.02$\\
DFN (N13N33N43) &  $\mathbf{94.13}$ & $\mathbf{72.92}$\\
DFN (N13N33N43N63N73) &  $93.97$ & $\mathbf{72.99}$\\
\hline
HighWay~\cite{SrivastavaGS15a} (19 layers) (2015) & $92.46$ & $-$ \\
HighWay~\cite{SrivastavaGS15a} (2015) & $92.40$ & $67.67$ \\
ResNet~\cite{HeZRS15b} (110 layers) (2015) & $93.57$ & $-$ \\
CMSC~\cite{LiaoC15a} (2015) & $93.13$ & $72.44$ \\
\hline
ALL-CNN~\cite{SpringenbergDBR14} (2014) & $92.75$ & $66.29$ \\
LSUV~\cite{MishkinM15} (19 layers) (2015) & $93.94$ &  $72.34$ \\
LSUV~\cite{MishkinM15} (maxout) (2015) & $\mathbf{94.16}$ &  $-$ \\
GPF~\cite{LeeGT15} (2015) & $93.95$ & $67.63$ \\
DSN~\cite{LeeXGZT15} (2015) & $92.03$ & $65.43$ \\
NiN~\cite{LinCY13} (2013) & $91.19$ & $64.32$ \\
Maxout~\cite{GoodfellowWMCB13} (2013) & $90.02$ & $65.46$ \\
MIN~\cite{ChangC15} (2015) & $93.25$ & $71.14$ \\
DNGO~\cite{SnoekRSKSSPPA15} (2015) & $93.63$  & $72.60$ \\
ELU~\cite{ClevertUH15} (2015) & $93.45$ & $\mathbf{75.72}$ \\
\hline
\multicolumn{3}{c}{Extreme data augmentation}\\
\hline
Large ALL-CNN~\cite{SpringenbergDBR14} (2014) & $95.59 $ & $-$ \\
Fractional MP~\cite{Graham14a} ($1$ test) (2014) & $95.50$ & $68.55$ \\
Fractional MP~\cite{Graham14a} ($12$ tests) (2014) & $\mathbf{96.53}$ & $73.61$ \\
SSCNN~\cite{Graham14} (2014) & $93.72$ & $75.70$  \\
\hline
\end{tabular}
  \vspace{-.3cm}
\end{table}

\section{Conclusion}
Deep fusion is an approach
that fuses not only the final representation
but also the intermediate representations of the base networks.
It is advantageous in
(1) Multi-scale representations can be learnt;
(2) The information flow is improved, and training a fused net composed
from a very deep base network and a shallow network
is less difficult than training the deep base network itself.
(3) The deep and shallow networks learning benefit from each other.
Experimental results show that our approach
achieves superior performance over ResNet
and Highway,
and competitive performance
compared to the state-of-the-arts.

\bibliographystyle{splncs}
\bibliography{deepfusion}

\end{document}